\documentclass[letterpaper]{article} 
\usepackage[submission]{aaai25}  
\usepackage{times}  
\usepackage{helvet}  
\usepackage{courier}  
\usepackage[hyphens]{url}  
\usepackage{graphicx} 
\usepackage{natbib}  
\usepackage{caption} 
\usepackage{algorithm}
\usepackage{algorithmic}

\usepackage{booktabs}
\usepackage{makecell}
\usepackage{amsmath}
\usepackage{amsfonts}
\usepackage{bm}
\usepackage{amssymb}
\usepackage{dsfont}
\usepackage{xcolor}
\DeclareMathOperator*{\argmax}{argmax}

\newcommand{\ourmethod}{\texttt{TaMatch}}
\usepackage[capitalize,noabbrev]{cleveref}
\usepackage{newfloat}
\usepackage{listings}
\DeclareCaptionStyle{ruled}{labelfont=normalfont,labelsep=colon,strut=off} 
\floatstyle{ruled}
\newfloat{listing}{tb}{lst}{}
\floatname{listing}{Listing}
\title{Towards the Mitigation of Confirmation Bias in Semi-supervised Learning: a Debiased Training Perspective}
\author{
    Yu Wang$^\ddag$,
    Yuxuan Yin$^\ddag$,
    Peng Li
}
\affiliations{
    University of California, Santa Barbara\\
    
    \{yu95, y\_yin, lip\}@ucsb.edu
}

\usepackage{bibentry}

\begin{document}

\maketitle
\def\thefootnote{$\ddag$}\footnotetext{Equal contribution.}

\begin{abstract}
Semi-supervised learning (SSL) commonly exhibits confirmation bias, where models disproportionately favor certain classes, leading to errors in predicted pseudo labels that accumulate under a self-training paradigm. Unlike supervised settings, which benefit from a rich, static data distribution, SSL inherently lacks mechanisms to correct this self-reinforced bias, necessitating debiased interventions at each training step. Although the generation of debiased pseudo labels has been extensively studied, their effective utilization remains underexplored. Our analysis indicates that data from biased classes should have a reduced influence on parameter updates, while more attention should be given to underrepresented classes.
To address these challenges, we introduce \ourmethod, a unified framework for debiased training in SSL. \ourmethod\ employs a scaling ratio derived from both a prior target distribution and the model's learning status to estimate and correct bias at each training step. This ratio adjusts the raw predictions on unlabeled data to produce debiased pseudo labels. In the utilization phase, these labels are differently weighted according to their predicted class, enhancing training equity and minimizing class bias. Additionally, \ourmethod\
dynamically adjust the target distribution in response to the model’s learning progress, facilitating robust handling of practical scenarios where the prior distribution is unknown. Empirical evaluations show that \ourmethod\ significantly outperforms existing state-of-the-art methods across a range of challenging image classification tasks, highlighting the critical importance of both the debiased generation and utilization of pseudo labels in SSL.
\end{abstract}

\section{Introduction}

The development of machine learning and artificial intelligence critically relies on availability of data. However, labeled data often remains scarce and expensive to obtain while unlabeled data is abundantly available. This disparity has propelled interest in semi-supervised learning (SSL), which leverages both labeled and unlabeled data. A prevalent SSL technique is pseudo labeling with a confidence threshold \citep{fixmatch, flexmatch, freematch}. This method involves utilizing the model under training to predict pseudo labels for unlabeled data;  predicted labels with a confidence level exceeding a certain threshold are then used to refine the model itself. The integration of this technique with other popular methods such as consistency regularization \citep{temporal, uda}, representation mixup \citep{mixmatch}, and distribution alignment\citep{remixmatch, softmatch} have been demonstrated great success.

Despite the widespread use of pseudo labeling, it is prone to ``confirmation bias"—a divergence defined as the discrepancy between the model’s predicted class probability of labels over the entire dataset ($\bm{p^{\text{model}}}$) and the true class distribution ($\bm{p^{\text{truth}}}$) during the training process. Confirmation bias is a very common phenomenon, which can stem from various sources, such as random initialization, different learned representation when using pre-trained models, the inherent learning difficulty across different classes, and batch variation. In SSL, unlike in fully-supervised learning where datasets are static, confirmation bias may cause the model to disproportionately favor certain classes, therefore continually refines its learning towards these classes, and eventually harms the model's generalization ability.

Recent approaches attempt to address confirmation bias either directly or indirectly. Methods such as de-biased SSL employ a separate head to decouple the generation and utilization of pseudo labels. Techniques like Distribution Alignment (DA) \citep{remixmatch} or Uniform Alignment (UA) \citep{softmatch} align the model’s predictions with a target distribution ($\bm{p^{\text{target}}}$), which is  either a known prior or a uniform distribution. This alignment is typically achieved by comparing the \emph{learning status}, typically an estimation of $\bm{p^{\text{model}}}$, with the target and adjusting model predictions before generating the actual pseudo labels. Other methods dynamically adjust the pseudo label acceptance threshold based on a predefined scheduler \citep{dash} or the model's learning status\citep{flexmatch, freematch}. Note that this line of works focus more on mitigating confirmation bias in the generation of pseudo labels.

Despite the efficacy of these strategies, there is a lack of exploration of \emph{joint debiased generation and utilization} of pseudo labels. We contend that even debiased pseudo labels should not contribute equally to the training process. Instead, they should be weighted according to the current learning status, reducing the learning rate for overrepresented classes to allow the model to concentrate on underrepresented data. This dual focus on both the debiased generation and utilization of pseudo labels is crucial for reducing confirmation bias in each iteration, leading to a more robust and stable learning process.


This paper proposes \ourmethod, a unified SSL framework incorporating debiased generation and utilization of pseudo labels to address confirmation bias. Leveraging a scaling ratio derived from a target distribution and the model's learning status for different classes, we first rescale the prediction of pseudo labels, and then assign different weights based on the corresponding class of their pseudo labels. Furthermore,  \ourmethod\ incorporates an exponential moving average (EMA) scheme to dynamically update the target distribution  in response to the model’s learning progress. 

We evaluate the proposed \ourmethod\ method on several challenging image classification tasks with bare supervision. \ourmethod\  achieves superior performance compared with existing state-of-the-art methods in both balanced and imbalanced settings, validating the necessity of debiased generation and utilization of unlabeled data in SSL.

\section{Preliminaries}\label{sec:prelim}
\subsection{SSL with Hard Pseudo Labeling and Consistency Regularization}
We consider a classification task involving $C$ classes and $d$-dimensional input features over a training dataset $\mathcal{D}:=\left\{(\bm{x}_{i}, \bm{y}_{i})\right\}_{i=1}^{N}$, where $\bm{x}\in \mathbb{R}^{d}$ denotes an input and $\bm{y}\in \mathbb{R}^{C}$ denotes the one-hot true label. A model $\mathcal{M}_{\bm{\theta}}$ which predicts the class label distribution $\bm{p}(\bm{y}|\bm{x})\in \mathbb{R}^{C}$ for an input $\bm{x}$ is trained by minimizing the following loss function over a mini-batch with $|\mathcal{B}|$ samples in one iteration:
\begin{equation}
    \begin{aligned}
        \mathcal{L} =
        \frac{1}{|\mathcal{B}|}\sum_{i \in \mathcal{B}}\left[\mathcal{H}(\bm{y_{i}}, \bm{p}(\bm{x_{i}}))\right]
    \end{aligned}
\end{equation}
where $\mathcal{H}(\cdot,\cdot)$ is a weighted cross-entropy function, and $\mathcal{B}$ is a uniformly sampled subset from an index set $\{1, \cdots, N\}$.

In a typical SSL setting, the training dataset is split into a labeled dataset: $\mathcal{D}^{l}:=\left\{(\bm{x}_{i}^{l},\bm{y}_{i}^{l})\right\}_{i=1}^{N_{l}}$ and a unlabeled dataset $\mathcal{D}^{u}:=\left\{\bm{x}_{i}^{u}\right\}_{i=1}^{N_{u}}$. Weak augmentation ($A^{w}(\cdot)$) and strong augmentation ($A^{s}(\cdot)$) \citep{temporal, uda} are also introduced to create diverse views of the data.

The training objective combines losses from labeled and unlabeled data: $\mathcal{L} = \mathcal{L}^l + \mathcal{L}^u$, where $\mathcal{L}^l$ is the cross-entropy loss on the labeled dataset. For the unlabeled data $\bm{x^{u}}$, the model first assigns pseudo labels $\bm{\hat{y}^{u}}$ by applying a high confidence threshold $\tau$ to  and the corresponding binary mask $m$ \citep{fixmatch} to the weakly augmented unlabeled data as in \cref{eq:pl}. This method ensures that only pseudo labels generated with a substantial confidence are included in model retraining.
\begin{align}
    \label{eq:pl}
    \bm{\hat{y}^{u}} &= \text{one-hot}(\text{argmax}(\bm{p}(\bm{y}|A^{w}(\bm{x^{u}})))), \\ 
    m &= \mathds{1}(\text{max}(\bm{p}(\bm{y}|\bm{x^{u}}))> \tau)
\end{align}

Then, the unlabeled loss, also known as consistency regularization, is defined by minimizing the cross-entropy loss between these pseudo labels and the predictions on the  strongly augmented versions of the same data:
\begin{equation}
    \begin{aligned}
        \mathcal{L}^{u} = 
        \frac{1}{|\mathcal{B}^{u}|}\sum_{i \in \mathcal{B}^{u}}\left[m_{i}\cdot \mathcal{H}(\bm{\hat{y}_{i}^{u}}, \bm{p}(\bm{y}|A^{s}(\bm{x_{i}^{u}})))\right]
    \end{aligned}
\end{equation}

In the rest of the paper, we abbreviate $\bm{p}(\bm{y}|A^{w}(\bm{x^{u}}))$ and $\bm{p}(\bm{y}|A^{s}(\bm{x^{u}}))$ by $\bm{p}^{w}$ and $\bm{p}^{s}$, respectively.  

\subsection{Confirmation Bias in SSL}
\paragraph{Definition:} In \ourmethod, confirmation bias is defined as the discrepancy between the model's expected marginal class predictions over the dataset, denoted by $\bm{p}^{\text{model}} := \mathbb{E}_{\mathcal{D}}[\bm{p}(\bm{y}|\bm{x})]$, and the true label distribution $\bm{p}^{\text{truth}}$. Additionally, we define a class $c$ as ``strong class" if $p^{\text{model}}(c) > p^{\text{truth}}(c)$, and as ``weak class" if $p^{\text{model}}(c) < p^{\text{truth}}(c)$. Furthermore, an unlabeled instance is classified as strong if it is predicted to belong to a strong class, and vice versa. 



\paragraph{Bias Amplification in SSL: }
The self-training nature of semi-supervised learning (SSL) can lead to the amplification of confirmation bias during the training process. Previous studies have primarily attributed this to the incorrect assignment of pseudo-labels, focusing their solutions on generating debiased pseudo-labels. However, our observations suggest that batch variation, inherent in mini-batch training, also plays a significant role in bias amplification. Unlike supervised settings, where the data distribution is static and batch variations can eventually be corrected, SSL is particularly vulnerable to these variations.

To illustrate this phenomenon, we present a simplified example that further motivates our proposal for the debiased utilization of pseudo-labels.

\paragraph{Motivating Example: }

We simplify the use of pseudo labels in $\mathcal{L}^{u}$ to a situation, where a categorical distribution use samples from itself to update its own parameter. Specifically, we consider a two-class categorical distribution parameterized by $\theta$: $\bm{p}(\theta):=(p_{1}, p_{2})$ where $p_{i}$ is the probability for the $i_{\text{th}}$ class:
\begin{align}
    p_{1} = \frac{1}{1 + e^{\theta}}, \quad p_{2} = \frac{e^{\theta}}{1 + e^{\theta}}
\end{align}

In each update step, $n$ samples are drawn from $\bm{p}(\theta)$ and the batch distribution $\tilde{\bm{p}}:=(\tilde{p}_{1}, \tilde{p}_{2})$ is estimated from these samples. The distribution parameter $\theta$ is then updated by minimizing the loss function that is defined as the KL divergence between $\tilde{\bm{p}}$ and $\bm{p}(\theta)$: $\mathcal{L}:=\mathbb{D}_{\text{KL}}\left[\tilde{\bm{p}} || \bm{p}(\theta)\right]$.


In this setup, we assume $\bm{p}^{\text{truth}}$ is $(0.5, 0.5)$. After steps of self-updating, bias amplification is defined as a situation where the KL divergence between $\bm{p}(\theta)$ and $\bm{p}^{\text{truth}}$ increases.

\begin{figure}
    \centering
    \includegraphics[width=0.93\linewidth]{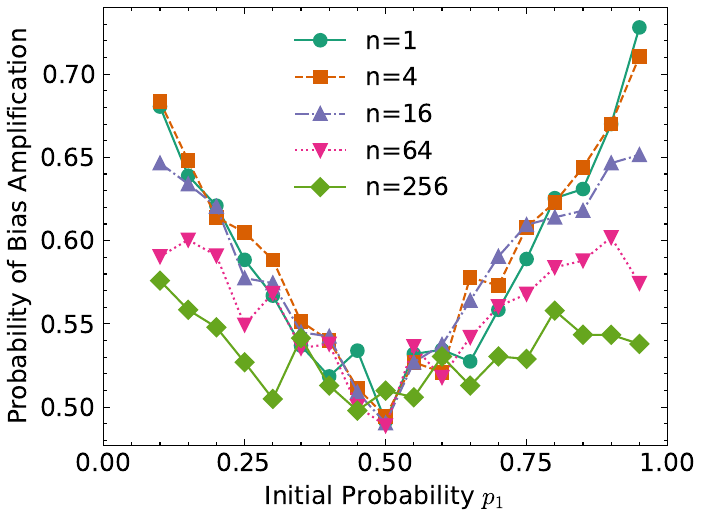}
    \caption{Probability of bias amplification with different initial $p_{1}$. $n$ is the number of samples drawn in each step.}
    \label{fig:toy}
\end{figure}

We run numerical simulation with different initial $p_{1}$ from 0.05 to 0.95. For each $p_{1}$, 1000 trajectories with 1000 update steps are simulated. We then average over all trajectories to calculate the probability of bias amplification and summarize them in \cref{fig:toy}.

There are two observations from our simulation. First, the probability of bias amplification increases as the initial distribution gets more biased. Second, small batch size leads to a clear increased bias amplification.
The second one is especially interesting as it indicates that when batch size is small, samples in the batch can potentially harm the learning process, even if they are indeed drawn from $\bm{p}$. This is not surprising as the batch variation can not be corrected when the distribution that generates samples is not static.

This motivates us that in SSL training, we should carefully utilize the predicted pseudo labels in mini-batches, even if they have been correctly assigned. Specifically, we propose that unlabeled data that is predicted to the strong class should be \emph{additionally} down-weighted to prevent the amplification of existing biases, and vice versa.


\section{Methodology}
We introduce a unified framework \ourmethod, which targets both the debiased generation and utilization of pseudo labels. Utilizing a scaling ratio derived from a ``target distribution" and the model's learning status, \ourmethod\ scales predictions on batched unlabeled data to produce debiased pseudo labels. In the next step of debiased pseudo label utilization,  pseudo labels accepted from the generation step are further weighted based on their predicted class so that they contribute differently to consistency loss to further reduce confirmation bias. 

\subsection{Overview of \ourmethod}

\begin{figure*}[t]
    \centering
    \includegraphics[width=0.95\linewidth]{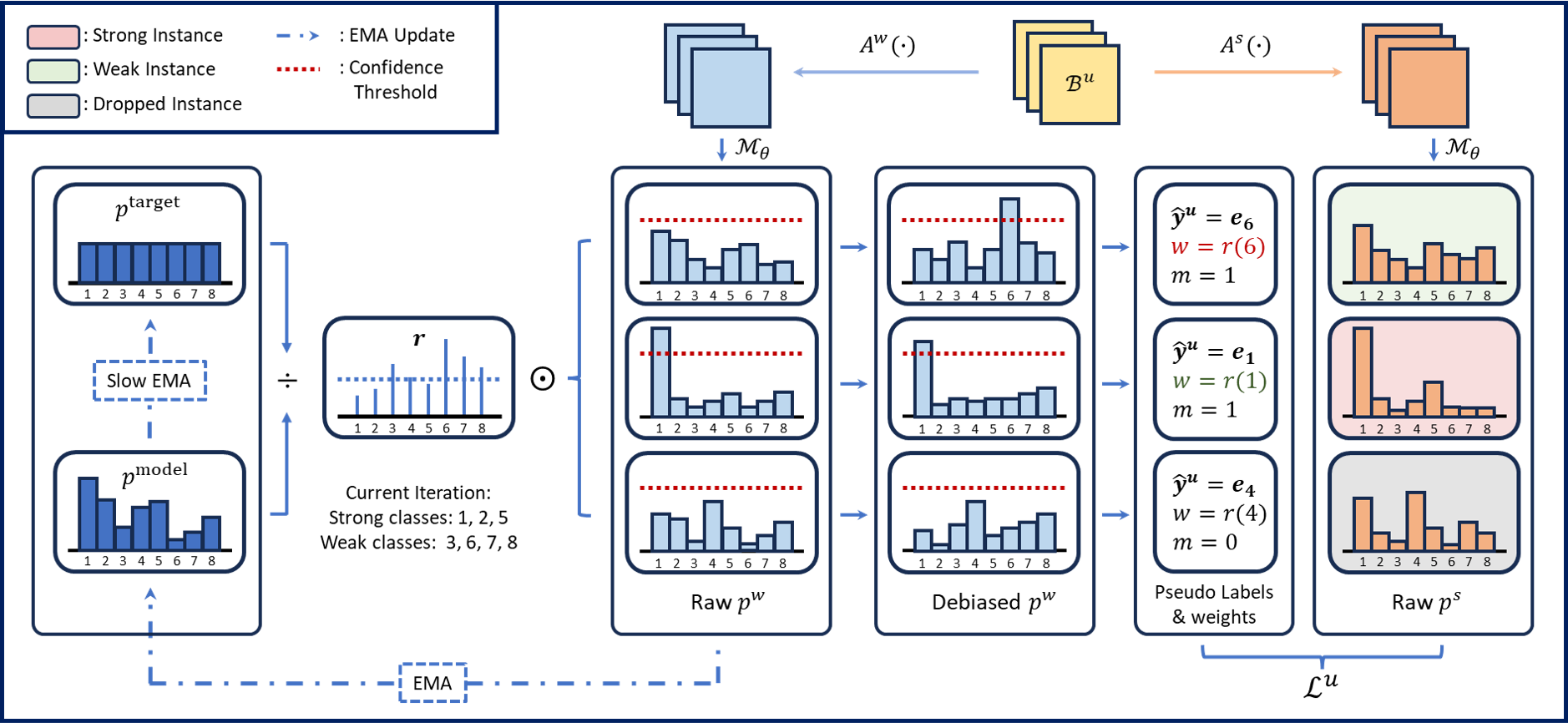}
    \caption{Overview of the \ourmethod\ framework with both debiased generation and utilization of pseudo labels.}
    \label{fig:overview}
\end{figure*}

\ourmethod\ involves three steps in each iteration to mitigate confirmation bias from pseudo-labels. First, a per-class scaling factor ($\bm{r}$) is calculated as the ratio between the target distribution and the current model's predicted class distributions. Given batches of weakly and strongly augmented unlabeled data, the model predicts their respective class probabilities ($\bm{p^{w}}$ and $\bm{p^{s}}$). $\bm{p^{w}}$ is then rescaled using the calculated scaling factors. Corresponding pseudo-labels and binary masks are derived from the rescaled $\bm{p^{w}}$ using a fixed high-confidence threshold ($\tau$). Meanwhile, instance weights are determined from the predicted pseudo-labels. Finally, the unlabeled data loss is computed for the batch based on the predicted pseudo-labels, binary masks, and corresponding weights. An illustrated overview is provided in \cref{fig:overview}.

\paragraph{Calculation of the Scaling Factor:} The per-class scaling factor is calculated as follows, where 
 $c$ is the class index:
\begin{align}
    \bm{r} = \frac{\bm{p^{\text{target}}}}{\bm{p^\text{model}}}, \ \ r(c)=\frac{p^{\text{target}}(c)}{p^{\text{model}}(c)}
\end{align}

Estimation of the true $\bm{p}^{\text{model}}$ is computationally infeasible \citep{remixmatch}. In practice, we approximate this using an exponential moving average (EMA) updated over the averaged $\bm{p^{w}}$ on the batch data:
\begin{align}
    \bm{p_{t}^{\text{model}}} = 
        \lambda^{\text{model}} \cdot \bm{p_{t-1}^{\text{model}}} + (1-\lambda^{\text{model}})\cdot \mathbb{E}_{\mathcal{B}^{u}}\left[\bm{p^{w}}\right]
\end{align}
where $\bm{p_{t}^{\text{model}}}$ is initialized as an uniform distribution, $\lambda^{\text{model}}\in [0, 1]$ is the momentum decay of EMA. We set $\lambda^{\text{model}}$ to 0.999, aligning to previous SSL approaches \citep{remixmatch, flexmatch, softmatch}.

\paragraph{Debiased Generation of Pseudo Labels:} To counteract the confirmation bias in a model, which tends to overpredict the probability of strong classes while underpredicting weaker ones \citep{remixmatch, softmatch}, we adjust the predicted probabilities $\bm{p^{w}}$ using the scaling factor, and normalize them to ensure the scaled probabilities sum to one:
\begin{align}
    \text{RE}(\bm{p^{w}}) :&= \text{Normalize}\left\{\bm{p^{w}}\odot \bm{r}\right\} \\
    \ &= \text{Normalize}\left\{\bm{p^{w}}\odot \frac{\bm{p^{\text{target}}}}{\bm{p^\text{model}}}\right\}.
\end{align}

We use $\text{RE}(\bm{p^{w}})$ to denote the rescaled prediction for weakly augmented data. The pseudo labels $\bm{\hat{y}^{u}}$ and the corresponding binary masks $m$ for the unlabeled data are then predicted based on the adjusted probabilities:

\begin{equation}
    \begin{aligned}
        \bm{\hat{y}^{u}} &= \text{one-hot}(\text{argmax}(\text{RE}(\bm{p^{w}})), \\  
        m &= \mathds{1}(\text{max}(\text{RE}(\bm{p^{w}}))> \tau)
    \end{aligned}
\end{equation}

\paragraph{Debiased Utilization of Pseudo Labels:} 
The debiased pseudo labels should contribute different to the training: indiscriminate reinforcement of predictions in favor of strong classes can exacerbate existing biases, which disrupts the learning process of the underrepresented classes. We modulate the influence of unlabeled instances based on their predicted classes by employing the scaling factor of the predicted class as the training weight of a given unlabeled data:
\begin{align}
    w = r(\argmax(\text{RE}(\bm{p^{w}}))
\end{align}

This ensures that reduced weights are assigned to strong instances and higher weights are assigned to weak ones. The final unlabeled loss in \ourmethod\ can be written as:
\begin{equation}
    \begin{aligned}
        \mathcal{L}^{u} = 
        \frac{1}{|\mathcal{B}^{u}|}\sum_{i \in |\mathcal{B}^{u}|}\left[w_{i}\cdot m_{i}\cdot \mathcal{H}(\bm{\hat{y}_{i}^{u}}, \bm{p^{s}})\right]
    \end{aligned}
\end{equation}

\subsection{Dynamically Updated Target Distribution}

The optimal target distribution, $\bm{p_{t}^{\text{target}}}$ should ideally be $p^{\text{truth}}$  which is typically not known a priori. Adopting a uniform target distribution during the early stages of model's training process  can encourage balanced learning over all class when the model is particularly vulnerable to the biases from random initialization and under-training. However, a strictly uniform target throughout the training might impede the model's ability to generalize to real-world, imbalanced data distributions.

To deal with these challenges, we have designed $\bm{p_{t}^{\text{target}}}$ as follows:
\begin{align}
    \bm{p_{t}^{\text{target}}} =
        \lambda^{\text{target}} \cdot \bm{p_{t-1}^{\text{target}}} + (1-\lambda^{\text{target}})\cdot \bm{p_{t}^{\text{model}}}
    \label{eq:p_target_update}
\end{align}

$\bm{p_{t}^{\text{target}}}$ is initialized as a uniform distribution. At the early training stage, more uniform $\bm{p^{\text{target}}}$ promotes balanced generation and utilization of pseudo labels across all classes. As the model becomes more accurate, the target distribution gradually follows the model's predicted probabilities. Eventually, the scaling factor $\bm{r}$ approaches $\bm{1}$, and no further adjustments to the model's prediction is applied.
\begin{table*}
    \centering
    \setlength{\tabcolsep}{1.6mm}{
    \small
    \caption{Error rate (\%)  on 3 semi-supervised image classification tasks in the USB benchmark. The mean and standard deviation of the error rate are reported across 3 random seeds}
    \label{tab:usb}
    \begin{tabular}{l | cc | cc | cc}
    \toprule
         \textbf{Dataset}&  \multicolumn{2}{c|}{\textbf{CIFAR-100}}  &  \multicolumn{2}{c|}{\textbf{STL-10}}  &  \multicolumn{2}{c}{\textbf{EuroSat}}  \\
         \# Labeled Data&  200&  400&  40&  100&  20&  40\\\midrule
         Fully Supervised&  8.3±0.08&  8.3±0.08&  -&  -&  0.94±0.03&  0.9±0.08\\
         Supervised&  35.88±0.36&  26.76±0.83&  19.0±2.9&  10.87±0.49&  26.49±1.6&  16.12±1.35\\\midrule
         Pseudo Labeling \citep{pseudo-labeling}&  33.99±0.95&  25.32±0.29&  19.14±1.30&  10.77±0.60&  25.46±1.36&  15.7±2.12\\
         Mean Teacher \citep{mean-teacher}&  35.47±0.40&  26.03±0.30&  18.67±1.69&  24.19±10.15&  26.83±1.46&  15.85±1.66\\
         $\Pi$-Model \citep{pi-model}&  36.06±0.15&  26.52±0.41&  42.76±15.94&  19.85±13.02&  21.82±1.22&  12.09±2.27\\
         VAT \cite{vat}&  31.49±1.33&  21.34±0.50&  18.45±1.47&  10.69±0.51&  26.16±0.96&  10.09±0.94\\
         MixMatch \citep{mixmatch}&  38.22±0.71&  26.72±0.72&  58.77±1.98&  36.74±1.24&  24.85±4.85&  17.28±2.67\\
         ReMixMatch \citep{remixmatch}&  22.21±2.21&  16.86±0.57&  13.08±3.34&  7.21±0.39&  5.05±1.05&  5.07±0.56\\
         AdaMatch \citep{adamatch}& 22.32±1.73& 16.66±0.62& 13.64±2.49& 7.62±1.90& 7.02±0.79& 4.75±1.10\\
         UDA \cite{uda}& 28.80±0.61& 19.00±0.79& 15.58±3.16& 7.65±1.11& 9.83±2.15& 6.22±1.36\\
         FixMatch \citep{fixmatch}& 29.60±0.90& 19.56±0.52& 16.15±1.89& 8.11±0.68& 13.44±3.53& 5.91±2.02\\
         FlexMatch \citep{flexmatch}& 26.76±1.12& 18.24±0.36& 14.40±3.11& 8.17±0.78& 5.17±0.57& 5.58±0.81\\
         Dash \citep{dash}& 30.61±0.98& 19.38±0.10& 16.22±5.95& 7.85±0.74& 11.19±0.90& 6.96±0.87\\
         CrMatch \citep{crmatch}& 25.70±1.75& 18.03±0.20& N/A & N/A& 13.24±1.69& 8.35±1.71\\
         CoMatch \citep{comatch}& 35.08±0.69& 25.35±0.50& 15.12±1.88& 9.56±1.35& 5.75±0.43& 4.81±1.05\\
         SimMatch \citep{simmatch}& 23.78±1.08& 17.06±0.78& \textbf{11.77±3.20}& 7.55±1.86& 7.66±0.60& 5.27±0.89\\
         FreeMatch \cite{freematch}& 21.40±0.30& \textbf{15.65±0.26}& 12.73±3.22& 8.52±0.53& 6.50±0.78& 5.78±0.51\\
         SoftMatch \citep{softmatch}& 22.67±1.32& 16.84±0.66& 13.55±3.16& 7.84±1.72& 5.75±0.62& 5.90±1.42\\
         DeFixMatch \citep{defixmatch}& 31.52±1.85& 21.12±1.74& 17.68±7.94& 7.94±1.31& 14.71±6.52& 3.72±0.79\\\midrule
         \ourmethod& \textbf{20.53±1.00}& 16.47±0.13& 14.11±1.65& \textbf{7.16±0.78}& \textbf{3.63±0.51}& \textbf{2.78±0.30}\\\bottomrule
    \end{tabular}
    }
\end{table*}

\subsection{Adaptive Weight Bound for Robust Training}
Confirmation bias can make the scaling factor $\bm{r}$ very different in order of magnitude, especially in the early stage of training. The large value range of $\bm{r}$ is not a problem for rescaling, but it impedes the effectiveness of the proposed weighting strategy in the training process: an extremely large weight of the tail class may incur unstable training dynamics, and a remarkably small weight of the head class can decrease the convergence speed. 

We propose to further incorporate the overall learning condition to set an adaptive bound $\left[r_{\text{min}}, r_{\text{max}}\right]$ for the scaling factor in weight computing:
\begin{equation}
    \left[r_{\text{min}}, r_{\text{max}}\right] = \left[1,1+\frac{\mathbb{D}_{\text{KL}}\left[\bm{p^{\text{model}}} || \bm{p^{\text{target}}}\right]}{\mathcal{H}(\bm{p^{\text{model}}})/C}\right]
\end{equation}

This adaptive range is hyper-parameter-free. It will converge to $[1,1]$ when $\bm{p^{\text{model}}} \to \bm{p^{\text{target}}}$, which means that $\mathbb{D}_{\text{KL}}\left[\bm{p^{\text{model}}} || \bm{p^{\text{target}}}\right] \to 0$. This property satisfies our design demand: the intensity of weighting should be proportional to the divergence of $\bm{p^{\text{model}}}$ to $\bm{p^{\text{target}}}$.

\section{Experiments}

\subsection{Balanced Image Classification}\label{sec:expr-usb}
\paragraph{Experimental Setup: } We evaluate \ourmethod\ on 3 datasets: CIFAR-100 \citep{cifar}, STL-10 \citep{stl}, and EuroSat \citep{eurosat}.  We conduct experiments on a sSL benchmark, named USB \citep{wang2022usb}. It has 2 settings for each image dataset, whose number of labeled data per class is 2/4 on CIFAR-100, 4/10 on STL-10, and 2/4 on EuroSat. The backbone neural network on each task is a pre-trained vision transformer (ViT) \citep{vit}. The optimizer is AdamW \citep{adamW} with a cosine scheduler. The batch size of both labeled data and unlabeled data is 8. The total training step is 204800, including 5120 steps of warm-up. All of the experiments are run with a single NVIDIA A100 GPU (80GB).
\begin{table}
    \centering
    \caption{Comparison of overall performance average across CIFAR-100, STL-10, and EuroSat in the USB Benchmark}
    \label{tab:usb-rank}
    \begin{tabular}{lcc}
    \toprule
         SSL Methods & \makecell[c]{Friedman\\ Rank}& \makecell[c]{Mean \\ Error Rate}\\
         \midrule
         FixMatch \citep{fixmatch}&  10.83& 15.46
\\
         FlexMatch \citep{flexmatch}&  7.17& 13.05
\\
         FreeMatch \citep{freematch}&  5.00& 11.76
\\
         SoftMatch \citep{softmatch}&  4.83& 12.09
\\ \midrule
         \ourmethod&  \textbf{2.00}& \textbf{10.78}\\
         \bottomrule
    \end{tabular}

\end{table}

\paragraph{\ourmethod\ Settings: } There are two hyper-parameters in \ourmethod: the pseudo label threshold $\tau$ and the momentum decay $\lambda^\text{model}$ of EMA updated $\bm{p^\text{model}}$. We set $\tau$ to 0.95, and $\lambda^\text{model}$ to 0.999, following recent semi-supervised approaches \citep{fixmatch, freematch}. In the balanced SSL settings, we do not update the target model, which means $\lambda^\text{target}=1$.

\begin{table*}[t]
    \centering
    \setlength{\tabcolsep}{2mm}{
\caption{Error rate (\%)  on the imbalanced CIFAR-10-LT in the USB benchmark. The mean and standard deviation of the error rate are reported across 3 random seeds}
\label{tab:imb}
    \begin{tabular}{l|cc|cc}
    \toprule
         \# Labeled, Unlabeled Data of Class 1& \multicolumn{2}{c|}{1500, 3000} &  \multicolumn{2}{c}{500, 4000}
\\ 
         Imbalance Ratio &  100&  150&  100& 150
\\ \midrule
         Fully Supervised&  26.17±0.33&  40.21±0.50&  53.37±0.88& 56.61±1.94
\\
         FixMatch \citep{fixmatch} &  23.31±0.78&  26.99±0.57&  27.59±1.71& 34.63±0.97
\\
         FreeMatch \citep{freematch} &  22.86±1.13&  26.71±0.63&  24.98±0.66& 32.23±1.02
\\ \midrule
         \ourmethod&  \textbf{21.84±0.60}&  \textbf{26.30±0.32}&  \textbf{24.56±0.51}& \textbf{32.08±0.57}
\\ \bottomrule
    \end{tabular}
    }
\end{table*}

\begin{figure*}[t]
    \centering
    \includegraphics[width=0.93\textwidth]{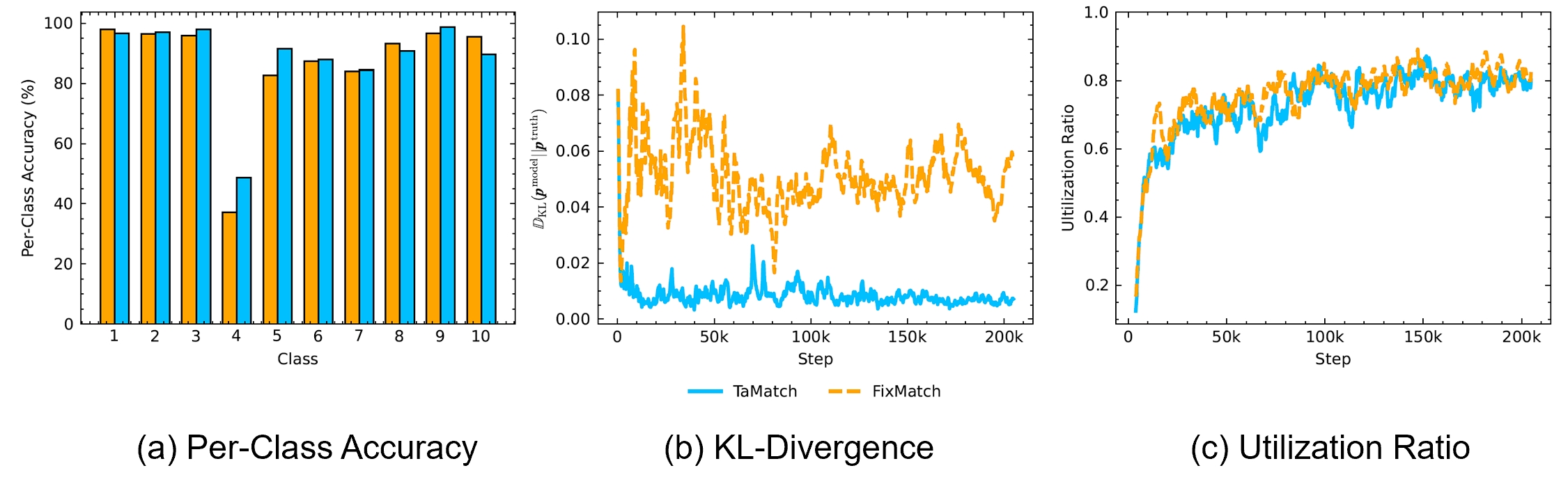}
    \caption{Comparison between \ourmethod\ and Fixmat on STL-10 with 40 labeled data. (a) Per-class accuracy; (b) KL divergence between $\bm{p}^{\text{model}}$ and $\bm{p}^{\text{truth}}$; (c) Utilization ratio of unlabeled data.}
    \label{fig:stl}
\end{figure*}

\paragraph{Baselines: } We compare \ourmethod\ with 17 SSL methods: Pseudo labeling \citep{pseudo-labeling}, Mean Teacher \citep{mean-teacher}, $\Pi$-model \citep{pi-model}, VAT \citep{vat}, MixMatch \citep{mixmatch}, ReMixMatch \citep{remixmatch},
AdaMatch \citep{adamatch}, UDA \citep{uda}, FixMatch \citep{fixmatch}, FlexMatch \citep{flexmatch}, Dash \citep{dash}, CRMatch \citep{crmatch}, CoMatch \citep{comatch}, SiMmatch \citep{simmatch}, FreeMatch \citep{freematch}, SoftMatch \citep{softmatch}, and DeFixMatch \citep{defixmatch}. Moreover, we add the performance of supervised learning is included for reference, using either training labeled data or the full dataset. All approaches are evaluated under the aforementioned settings, and their hyper-parameters are set to the values presented in their original paper. We present the results of these baselines provided by the USB benchmark.

We report the mean and the standard deviation of the test error rate (\%) across 3 random seeds. Moreover, we use the Friedman rank and the average error rate of all semi-supervised image classification tasks for a fair comparison.

\paragraph{Results: } As shown in \cref{tab:usb} and \cref{tab:usb-rank}, \ourmethod\ matches or surpasses the state-of-the-art SSL baselines. It achieves the highest Friedman rank and the lowest mean error rate, demonstrating its superior effectiveness. Additionally, \ourmethod's standard deviation is consistently smaller than that of FixMatch, indicating its training robustness.

\subsection{Long-Tailed Image Classification} \label{sec:expr-imb}

One important cause of confirmation bias is from the inherently imbalanced dataset. Consequently, we further validate the efficacy of \ourmethod\ in mitigating confirmation bias for long-tailed image classification tasks.

\begin{table}
\centering
\caption{Ablation study of \ourmethod\ on CIFAR-100. The mean  and standard deviation of the error rate are reported across 3 random seeds.}
\label{tab:ablation-usb}
    \begin{tabular}{l|cc}
    \toprule
         \# Labeled Data&  200&  400  \\\midrule
         FixMatch\citep{fixmatch}&  29.60±0.90 & 19.56±0.52
\\ \midrule
         \ourmethod\ w/o Rescaling&  23.27±0.34& 16.91±0.85
\\
         \ourmethod\ w/o Weighting&  22.38±0.93& 16.62±0.03
\\
         \ourmethod\ &  20.53±1.00& 16.47±0.13
\\\bottomrule
    \end{tabular} 
\end{table}

\begin{table*}[t]
    \centering
    \caption{Ablation study of \ourmethod\ on imbalanced CIFAR-10-LT. The mean  and standard deviation of the error rate are reported across 3 random seeds.}
    \label{tab:imb-ablation}
    \begin{tabular}{l|cc|cc}
    \toprule
         \# Labeled, Unlabeled Data of Class 1& \multicolumn{2}{c|}{1500, 3000} &  \multicolumn{2}{c}{500, 4000}
\\ 
         Imbalance Ratio &  100&  150&  100& 150
\\ \midrule
         FixMatch \citep{fixmatch}&  23.31±0.78&  26.99±0.57&  27.59±1.71& 34.63±0.97
\\\midrule
         \ourmethod\ w/o Clipping&  21.86±0.38&  26.81±0.42&  24.32±1.72& 35.41±1.45
\\
         \ourmethod\ w/o EMA Target&  22.52±0.21&  26.40±0.48&  27.44±1.71& 32.38±2.21\\ 
         \ourmethod&  21.84±0.60&  26.30±0.32&  24.56±0.51& 32.08±0.57
\\ \bottomrule
    \end{tabular}
\end{table*}
\begin{figure}[t]
    \centering
    \includegraphics[width=0.33\textwidth]{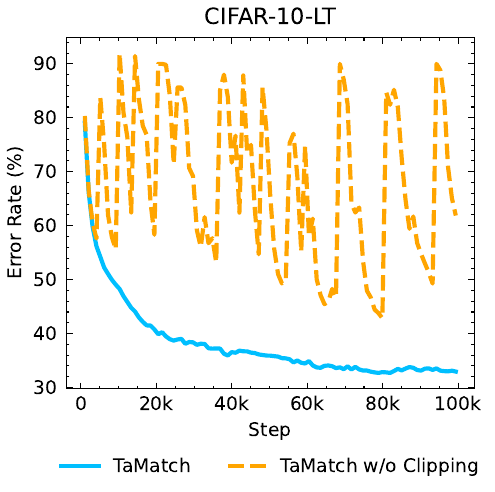}
    \caption{Error rate during training of \ourmethod\ on the imbalanced CIFAR-10-LT with $N_l^{(1)}=500$, $N_u^{(1)}=4000$, and $\gamma=150$.}
    \label{fig:imb-clipping}
\end{figure}

\paragraph{Experimental Setup: } 
We demonstrate the effectiveness of \ourmethod\ on a challenging semi-supervised image classification task with imbalanced data distribution. Experiments are conducted on the USB benchmark. We consider various imbalance settings of CIFAR-10-LT \citep{cifar}. The number of labeled training examples $N_l^{(i)}$ and unlabeled training examples $N_u^{(i)}$ in class $i$ is controlled by an imbalance ratio $\gamma$: $N_l^{(i)} = N_l^{(1)} \cdot \gamma ^{-\frac{i-1}{9}}$ and $N_u^{(i)} = N_u^{(1)} \cdot \gamma ^{-\frac{i-1}{9}}$. $(N_l^{(1)}, N_u^{(1)})$ is set to $(1500, 3000)$ or $(500, 4000)$, and $\gamma$ is set to 100 or 150. The backbone neural network on all settings is a Wide ResNet-28-2 \citep{wide-resnet}. The optimizer is SGD with a cosine scheduler. The batch size of labeled data and unlabeled data are 64 and 128, respectively. The total training step is 262144 without warm-up. 

\paragraph{\ourmethod\ Settings: } In the imbalanced dataset, since the ground truth class marginal distribution is unknown and setting a uniform $\bm{p^{\text{target}}}$ can potentially prevents the model from generalization. We use \cref{eq:p_target_update} to udpate the $\bm{p^{\text{target}}}$ with $\lambda^\text{ref}$ to 0.99999. We set other hyperparameters to be $\tau$ to 0.95, $\lambda^\text{model}$ to 0.999, following baseline methods \citep{fixmatch, softmatch, freematch}.



\paragraph{Baselines: } We compare \ourmethod\ to compatible semi-supervised learning methods: FixMatch \citep{fixmatch} and FreeMatch \citep{freematch}. The supervised learning method is added for reference. All approaches follow the setting as above.

We report the mean and the standard deviation of the test error rate (\%) across 3 random seeds.

\paragraph{Results: } \cref{tab:imb} shows the results on CIFAR-10-LT. \ourmethod\ outperforms the SSL baselines on all imbalance settings in both the mean and std of error rate, demonstrating its superior efficacy.

\subsection{Qualitative Analysis}
We evaluate how \ourmethod\ mitigates training bias by examining the training process. We conduct experiments on STL-10 with 4 labels per class, comparing \ourmethod\ with a FixMatch counterpart that does not employ any debiasing strategy. We compare per-class accuracy, the KL-Divergence between the $\bm{p^{\text{model}}}$ and the ground truth (Uniform) distribution, and the utilization ratio, which is defined as the number of pseudo labels with a non-zero binary mask over the batch size. These comparisons are plotted in \cref{fig:stl}.

\ourmethod\ demonstrates a clearly reduced KL-Divergence throughout the training process, whereas FixMatch shows large oscillations. Notably, our utilization ratio is roughly the same as FixMatch, indicating that the benefits of \ourmethod\ do not come from introducing more unlabeled data, but from better utilization. The final per-class accuracy also supports that \ourmethod\ effectively balances learning towards the weaker class 4, whose accuracy is notably lower compared to other classes in FixMatch.

\subsection{Ablation Studies}
We demonstrate the effectiveness of the proposed techniques in \ourmethod: reweighting, rescaling, and EMA updating of the marginal distribution target model.

\subsubsection{5.4.1 Effectiveness of Reweighting and Rescaling}
\paragraph{Experimental Setup: } We validate the effectiveness of reweighting and rescaling of \ourmethod\ on the balanced CIFAR-100 dataset, with access to 2 or 4 randomly sampled labeled data per class. The experimental settings are the same as those discussed in \cref{sec:expr-usb}. 

\paragraph{Results: } As presented in \cref{tab:ablation-usb}, while the removal of either proposed technique from \ourmethod\ still outperforms the baseline FixMatch, it yields an increase in error rate, especially in the case of fewer labels. This result verifies the essentials of rescaling and reweighting for \ourmethod's overall performance.

\subsubsection{5.4.2 Effectiveness of Target Updating and Clipping}
\paragraph{Experimental Setup: } We conduct experiment on the imbalanced CIFAR-10-LT dataset, with 4 different settings presented in \cref{sec:expr-imb}. 

\paragraph{Results: } We show the error rate on CIFAR-10-LT in \cref{tab:imb-ablation}. The removal of the EMA updated target model degrades the performance, especially in $\gamma=100$. In contrast, clipping is more crucial in $\gamma=150$ cases. \cref{fig:imb-clipping} illustrates the significance of clipping for \ourmethod\ in the severe imbalanced dataset.

\section{Related Work}
Pseudo-labeling \citep{pseudo-labeling} and consistency regularization \citep{temporal} are two fundamental techniques in semi-supervised learning (SSL). Pseudo-labeling generates artificial labels for unlabeled data to enable self-training, while consistency regularization encourages consistent predictions for similar input data points. A significant body of work has focused on improving and extending these core methods.

FixMatch \citep{fixmatch} employs a fixed, high-confidence threshold to generate pseudo-labels, ensuring only high-quality unlabeled data is included in the training. Other approaches focus on adjusting the threshold to adapt to the model's learning curve. Dash \citep{dash} uses a predefined scheduler to adjust the threshold and has proven the convergence property. FlexMatch \citep{flexmatch} dynamically adjusts the threshold for different classes by estimating the learning effect of each class. FreeMatch \citep{freematch} introduces both global and per-class threshold adjustments based on an estimation of the model's learning status. Distribution Alignment (DA) \citep{remixmatch} encourages the model's predictions to align with the ground truth marginal label distribution by directly adjusting the raw predicted class probability. A similar technique, Uniform Alignment (UA) \citep{softmatch}, is utilized to encourage a balanced quantity between different classes.

Another line of works focus on assigning different weights to unlabeled data. \citep{ren2020unlabeled} uses the influence function to weight unlabeled instances by their calculated "importance" on the validation dataset. \citep{iscen2019label} targets the transductive setting and uses uncertainty and an estimation of class population to derive per-sample and per-class weights. SoftMatch \citep{softmatch} addresses the quality-quantity tradeoff in SSL by introducing a truncated Gaussian mask, which assigns weights to unlabeled data points based on the corresponding confidence.

Regarding directly addressing the confirmation bias issue, \citep{chen2022debiased} proposes debiasing pseudo-labels using counterfactual reasoning, while \citep{wang2022debiased} introduces a separate head into the network architecture to decouple the generation and utilization of pseudo-labels.

\section{Limitation and Future Work}\label{sec:limitation}
Several limitations exist in the current implementation of \ourmethod. First, a fixed high confidence threshold, directly adopted from FixMatch \citep{fixmatch}, can reduce the number of pseudo labels and slow down the overall training process, as discussed in \citep{freematch, softmatch, flexmatch}. Further analysis on the impact of including more (potentially low-quality) data on confirmation bias, and how to appropriately leverage this data to mitigate bias, remains an intriguing area for future research. Second, there is a lack of theoretical analysis on how to optimally adjust the model's raw predictions and apply weighting. Currently, the rescaling and reweighting based on the model's learning status are largely qualitative. Conducting quantitative analysis towards optimal debiased generation and utilization of pseudo labels will be a significant and impactful direction for future work.

\bibliography{aaai25}

\appendix
\onecolumn

\section{Motivating Example Setup}
The example targets a scenario where a two-class categorical distribution use samples drawn from itself to update its parameters across iterations. 

Let $\bm{p}(\theta):=(p_{1}, p_{2})$ denotes the distribution parameterized by $\theta$ where:
\begin{align}
    \label{eq:p_cat}
    p_{1}(\theta) = \frac{1}{1 + e^{\theta}}, \quad p_{2}(\theta) = \frac{e^{\theta}}{1 + e^{\theta}},
\end{align}
denotes the probability of class 1 and class 2, respectively. This parameterisation is commonly adopted in classification tasks where the soft max score of classes is used to predict the labels.

In each iteration, a batch of $n$ samples $X$ is drawn from the current distribution and the batch distribution $\bm{\tilde{p}}:=(\tilde{p})$ is calculated as:
\begin{align}
    \tilde{p}_{i} = \frac{1}{n}\sum_{j=1}^{n} \mathbf{1}(X_{j} = i).
\end{align}

The parameter $\theta$ is updated to minimize the loss function defined as the Kullback–Leibler (KL) divergence from $\bm{\tilde{p}}$ to $\bm{p}(\theta)$:
\begin{align}
\mathcal{L(\theta)}:=\mathbb{D}_{\text{KL}}\left[\tilde{\bm{p}} || \bm{p}(\theta)\right] = \sum_{i=1}^{n}\tilde{p}_{i}\cdot \left[\log{\frac{\tilde{p}_{i}}{p_i(\theta)}}\right]
\end{align}

With \cref{eq:p_cat}, this loss function can be further simplified to:
\begin{align}
    \mathcal{L(\theta)} &= \tilde{p}_{1}\cdot \log{\tilde{p}_{1}} + \tilde{p}_{2}\cdot \log{\tilde{p}_{2}} + \tilde{p}_{1}\cdot \log{(1 + e^{\theta})} - \tilde{p}_{2}\cdot \log{\frac{e^{\theta}}{1 + e^{\theta}}}\\
    &= \log{(1 + e^{\theta})} - \tilde{p}_{2}\cdot \theta + \tilde{p}_{1}\cdot \log{\tilde{p}_{1}} + \tilde{p}_{2}\cdot \log{\tilde{p}_{2}},
\end{align}

and the gradient to update $\theta$ can be derived as:
\begin{align}
    \frac{d\mathcal{L}(\theta)}{d\theta} &= \frac{e^\theta}{1 + e^{\theta}} - \tilde{p}_{2}\\
    &=\tilde{p}_{1} - p_{1}(\theta).
\end{align}

With gradient descent method, we have:
\begin{align}
    \label{eq: grad}
    \theta^{'} = \theta - \eta \cdot (\tilde{p}_{1} - p_{1}(\theta)),
\end{align}
where $\eta$ is the update step size. 

When $n$ is large, the stochastic gradient can be approximated with a Gaussian noise and update of $\theta$ degrades to Brownian motion around the initial $\theta$. However, when $n$ is small, which is the small batch size case we are interested in, analytical solution for $\theta$ under steps of update becomes intractable. Thus, we use \cref{eq: grad} to run numerical solution for a total of 20 initial probabilities from 0.05 to 0.95. For each initial probability, the setting is as below:
\begin{itemize}
    \item Number of trajectories: 1000 
    \item Number of steps per trajectory: 1000
    \item Step size: 1
\end{itemize}
\section{Open Source Code}
We have made our code openly available at: \url{https://anonymous.4open.science/r/TaMatch-Official-CDDF/}.
\section{Hyper-parameters Settings} \label{sec:hyper-params}

We adopt hyperparameters following the default setup for FixMatch in the USB benchmark \cite{wang2022usb} and did not run additional hyper parameter search. We report these settings for both balanced and imbalanced SSL tasks below in \cref{tab:usb-hyper} and \cref{tab:usb-imb-hyper}, respectively.

\begin{table*}[h]
    \centering
\caption{Hyper-parameters of balanced SSL tasks in the USB benchmark}
\label{tab:usb-hyper}
    \setlength{\tabcolsep}{1.6mm}{
    \begin{tabular}{lccc}
    \toprule
         Dataset&  CIFAR-100&  STL-10&  EuroSat\\
         \midrule
         Image Size&  32&  96&  32\\
         Model&  ViT-S-P2-32&  ViT-B-P16-96&  ViT-S-P2-32\\
         Weight Decay&  \multicolumn{3}{c}{5e-4} \\
         Labeled Batch size&  \multicolumn{3}{c}{8}  \\
         Unlabeled Batch size&  \multicolumn{3}{c}{8}  \\
         Learning Rate&  5e-4&  1e-4&  5e-5\\
         Layer Decay Rate&  0.5&  0.95&  1.0\\
         Scheduler&  \multicolumn{3}{c}{$\eta=\eta_0 \cos(\frac{7\pi k}{16K})$}  \\
         Model EMA Momentum&  \multicolumn{3}{c}{0.0}  \\
 Prediction EMA Momentum& \multicolumn{3}{c}{0.999} \\
 Weak Augmentation& \multicolumn{3}{c}{Random Crop, Random Horizontal Flip} \\
 Strong Augmentation& \multicolumn{3}{c}{RandAugment \cite{randaugment}} \\
 \bottomrule
    \end{tabular}
    }
    
\end{table*}

\begin{table*}[h]
    \centering
\caption{Hyper-parameters of imbalanced SSL tasks in the USB benchmark}
\label{tab:usb-imb-hyper}
    \setlength{\tabcolsep}{2.9mm}{
    \begin{tabular}{lc}
    \toprule
         Dataset&  CIFAR-10-LT\\
         \midrule
         Image Size&  32\\
         Model&  Wide ResNet-28-2\\
         Weight Decay&  0.03\\
         Labeled Batch size&  64\\
         Unlabeled Batch size&  128\\
         Learning Rate&  5e-4\\
         Layer Decay Rate&  1.0\\
         Scheduler&  $\eta=\eta_0 \cos(\frac{7\pi k}{16K})$\\
         Model EMA Momentum&  0.0\\
 Prediction EMA Momentum& 0.999\\
 Weak Augmentation& Random Crop, Random Horizontal Flip\\
 Strong Augmentation& RandAugment \cite{randaugment}\\
 \bottomrule
    \end{tabular}
    }
    
\end{table*}

\end{document}